\documentclass[letterpaper, 10 pt, conference]{ieeeconf}  % Comment this line out if you need a4paper

\IEEEoverridecommandlockouts                              % This command is only needed if 
\usepackage[table, dvipsnames]{xcolor}
\usepackage[pdftex]{graphicx} % graphics package
\usepackage{graphicx}
\usepackage[lofdepth,lotdepth,caption=false]{subfig} % subplot package
\graphicspath{{figures/}} % declare the path where graphic files are
\DeclareGraphicsExtensions{.pdf,.jpg,.png} % grafic files extensions
\usepackage{amsmath} % math package for split
\usepackage{amsfonts} % math package for fonts
\usepackage{mathtools} % math package for tools like prescript
\usepackage{tikz} % figure package
\usetikzlibrary{shapes,arrows,fit,calc,positioning,automata,backgrounds}
\usepackage{color} % colors package
\usepackage{multirow} % multiple rows in table
\usepackage{hhline} % separator in table
\usepackage[bookmarks=false]{hyperref} % hyperlink package
\usepackage{enumerate} % enumerate package
\usepackage{epstopdf} % package to include eps. files
\usepackage{dsfont}
\usepackage{siunitx} % package for international system units
\usepackage{cite}   % for compact citations, i.e. [1, 2]
\usepackage{diagbox}    % for making slash cell in a table
\usepackage{soul}

\colorlet{Green1}{green!90!}
\colorlet{Green2}{green!60!}
\colorlet{Green3}{green!40!}
\colorlet{Green4}{green!20!}
\colorlet{Green5}{green!10!}

\definecolor{Bookcolor}{HTML}{00F9DE}

\makeatletter
\def\@citex[#1]#2{\leavevmode
\let\@citea\@empty
\@cite{\@for\@citeb:=#2\do
{\@citea\def\@citea{,\penalty\@m\ }%
\edef\@citeb{\expandafter\@firstofone\@citeb\@empty}%
\if@filesw\immediate\write\@auxout{\string\citation{\@citeb}}\fi
\@ifundefined{b@\@citeb}{\hbox{\reset@font\bfseries ?}%
\G@refundefinedtrue
\@latex@warning
{Citation `\@citeb' on page \thepage \space undefined}}%
{\@cite@ofmt{\csname b@\@citeb\endcsname}}}}{#1}}
\makeatother

\begin{document}

\title{Curiosity-Driven Reinforcement Learning based Low-Level Flight Control}

% use a multiple column layout for up to three different affiliations
\author{Amir Ramezani Dooraki and Alexandros Iosifidis
% <-this % stops a space
\thanks{A. Ramezani Dooraki and A. Iosifidis are with the Department of Electrical and Computer Engineering, Aarhus University,
        8000 Aarhus C, Denmark
        {\tt\small \{amir, ai\} at ece.au.dk}}%
}


%%\author{\IEEEauthorblockN{Michael Shell\IEEEauthorrefmark{1},
%%Homer Simpson\IEEEauthorrefmark{2},
%%James Kirk\IEEEauthorrefmark{3}, 
%%Montgomery Scott\IEEEauthorrefmark{3} and
%%Eldon Tyrell\IEEEauthorrefmark{4}}
%%\IEEEauthorblockA{\IEEEauthorrefmark{1}School of Electrical and Computer Engineering\\
%%Georgia Institute of Technology,
%%Atlanta, Georgia 30332--0250\\ Email: see http://www.michaelshell.org/contact.html}
%%\IEEEauthorblockA{\IEEEauthorrefmark{2}Twentieth Century Fox, Springfield, USA\\
%%Email: homer@thesimpsons.com}
%%\IEEEauthorblockA{\IEEEauthorrefmark{3}Starfleet Academy, San Francisco, California 96678-2391\\
%%Telephone: (800) 555--1212, Fax: (888) 555--1212}
%%\IEEEauthorblockA{\IEEEauthorrefmark{4}Tyrell Inc., 123 Replicant Street, Los Angeles, California 90210--4321}}


\maketitle

% insert page header and footer here for IEEE PDF Compliant
%\thispagestyle{fancy}
%\fancyhead{}
%\lhead{}
%\lfoot{}
%\cfoot{}
%\rfoot{}
%\renewcommand{\headrulewidth}{0pt}
%\renewcommand{\footrulewidth}{0pt}

\begin{abstract}
Curiosity is one of the main motives in many of the natural creatures with measurable levels of intelligence for exploration and, as a result, more efficient learning. It makes it possible for humans and many animals to explore efficiently by searching for being in states that make them surprised with the goal of learning more about what they do not know. As a result, while being curious, they learn better. In the machine learning literature, curiosity is mostly combined with reinforcement learning-based algorithms as an intrinsic reward. 
This work proposes an algorithm based on the drive of curiosity for autonomous learning to control by generating proper motor speeds from odometry data. The quad-copter controlled by our proposed algorithm can pass through obstacles while controlling the Yaw direction of the quad-copter toward the desired location. To achieve that, we also propose a new curiosity approach based on prediction error. 
%We implement our algorithm in Python and test it with the Gazebo simulator controlled through our Python interface and Robot Operating System. 
We ran tests using on-policy, off-policy, on-policy plus curiosity, and the proposed algorithm and visualized the effect of curiosity in evolving exploration patterns. Results show the capability of the proposed algorithm to learn optimal policy and maximize reward where other algorithms fail to do so. 
%Finally, we provide discussion and conclude our paper.   
\end{abstract}
\IEEEpeerreviewmaketitle

%\bstctlcite{IEEEexample:BSTcontrol}

\section{Introduction}
Humans and intelligent creatures can learn in different ways; among them is learning by experience. Further, they use a spectrum of motivations: some are triggered internally (intrinsic motivations), and some are triggered externally and by the environment (extrinsic motivations). An intelligent creature learns to act in the direction of responding to its motivations. 
Observing this paradigm in nature, machine learning and control communities created the framework of the Markov Decision Process (MDP) and Reinforcement Learning (RL) algorithms to replicate this optimization process in robots and machines. Further, several computational models of intrinsic motivations, such as curiosity, have been implemented in the past decades.
\par
At the same time, considering the advances in computer hardware, different intelligent algorithms for autonomous control of ground, aerial, underwater, surface, and legged robots have been created during the past decade. Specifically, these advancements were significant for multi-copter drones where the weight of the whole robot is significant in terms of its ability to fly and maneuver capabilities in three axes. Nowadays, it is possible to see the application of Unmanned Aerial Vehicles (UAVs) such as quad-copters in several areas. Autonomous inspection, search and rescue missions, and navigation in unknown environments are examples of high-level control where an algorithm takes high-level decisions and passes it to a low-level controller for execution using robots' actuators. Autonomous learning of aggressive maneuvers, drone racing, and fault-tolerant control are examples of low-level flight controllers where the algorithm directly controls the actuators.
\par
Combining the framework of learning-based algorithms, such as reinforcement learning and deep reinforcement learning, with UAVs pushed their autonomous control to new frontiers, allowing them to autonomously learn to control both in high-level and low-level state spaces. 
\par
This paper proposes a new reinforcement learning-based low-level flight controller that learns by parameterized intrinsic (a computational model of curiosity) and extrinsic (external immediate and auxiliary rewards) motivations to directly control the quad-copter's motor speeds and flies toward the desired position while avoiding obstacles. Our contributions are summarized as follows:
\begin{itemize}
  \item We propose a new approach for learning low-level flight policy using parameterized curiosity module.
  \item In our proposed approach, we consider both passing through obstacles and controlling the Yaw direction towards the desired location.
  \item To achieve the mentioned contributions, we propose a new approach for calculating the curiosity reward based on the prediction error.
\end{itemize}
\par
In the rest of this paper, first, we discuss the related literature and the difference between our work and other related works. Next, we describe the proposed methodology including the reinforcement learning approach, the curiosity module, the simulation environment, and the visualization of the curiosity effect. Then, we describe the experimental evaluation and provide results along with a discussion about important matters related to our work. Finally, we provide concluding remarks.

\section{Literature Review}
In the literature on unmanned aerial vehicles' control using reinforcement learning, a wide range of works exists that can be divided into two main groups, namely high-level and low-level control. By low-level control, we refer to the direct control of motors by providing their actual angular velocity or by providing thrust, roll, pitch, and yaw. Low-level control works include \cite{RAMEZANIDOORAKI2021103671} where RL is used to learn for direct control of UAV motors speeds from odometry, \cite{Hwangbo_2017} where a combination of RL and PD is used to train the controller, \cite{MolchanovRL_LowLevel} where RL used to learn more general policies for low-level quad-copter control, and \cite{ChenHuanPi_LowLevel} where RL is used for control and tracking of a trajectory. By high-level control, we refer to a trajectory of waypoints or attitudes generated by the controller. High-level control works include \cite{machines10070500} where an end-to-end approach using deep RL is used to learn to control three axes of quad-copter when RGB-D image is provided as input.
\par
Curiosity as an intrinsic motive is observed widely in the literature. In some early works, such as \cite{intrinsicmotivation} and \cite{GOTTLIEB2013585}, the authors defined a framework of intrinsic motivation and curiosity as one such motivation. In recent years, and considering the new computational capabilities offered by advancements in hardware systems, more realistic computational models of curiosity have been researched and developed in the literature. As a result, different types of curiosity methods, such as Information Theoretic based \cite{DBLP:journals/corr/HouthooftCDSTA16}, Prediction based (e.g., surprisal \cite{DBLP:journals/corr/PathakAED17}), and Count based \cite{10.5555/3305890.3305962}, have been proposed. For example, a module called intrinsic curiosity module (ICM) is used in \cite{DBLP:journals/corr/PathakAED17} to predict the future states and actions that need to be taken to reach those states, and considered the prediction error between the actual future states and predicted states as the curiosity reward. Further, there are works such as \cite{10.5555/3305890.3305962} which considered curiosity as a measure to count how many times a state is visited during the agent's lifetime. Curiosity has been used to achieve specific qualities in robot actions. For example, \cite{LearningGentleObject} used curiosity to achieve gentle touch while grasping objects. In other literature, it used to increase the performance of well-known objectives in robotics. For example, in \cite{motionplanninghumanoid} curiosity is used for motion planning of humanoid robots, in \cite{InnateMotivationRobotSwarms} for control of swarm of robots, and in \cite{CuriosityRobotNavigation} for robot navigation.
\par
Curiosity in the low-level control of the quad-copter is an area that is much less researched in the literature. In some works, curiosity is defined as a measure of difference rather than surprise or novelty. For example, \cite{sun2022aggressive} defines curiosity as the difference between states observed by a policy at two different times. 
%\par
In this work, curiosity is defined as a measure of novelty and a parameterized function that gradually learns the previously visited states, loses its interest in them, and constantly searches for novel states.

% \begin{figure*}[th]
% \includegraphics[scale=0.25]{sections/images/errors_all_idealistic.png}
% \caption{This figure shows the position and attitude errors on X, Y, and Z for position and Roll, Pitch, and Yaw for the attitude (all errors are related to tests in an Idealistic environment).}
% \label{position_error_image}
% \end{figure*}
\section{Methodology}
\subsection{Main Algorithm}
\begin{figure*}[t]
\centering
\includegraphics[width=0.75\textwidth]{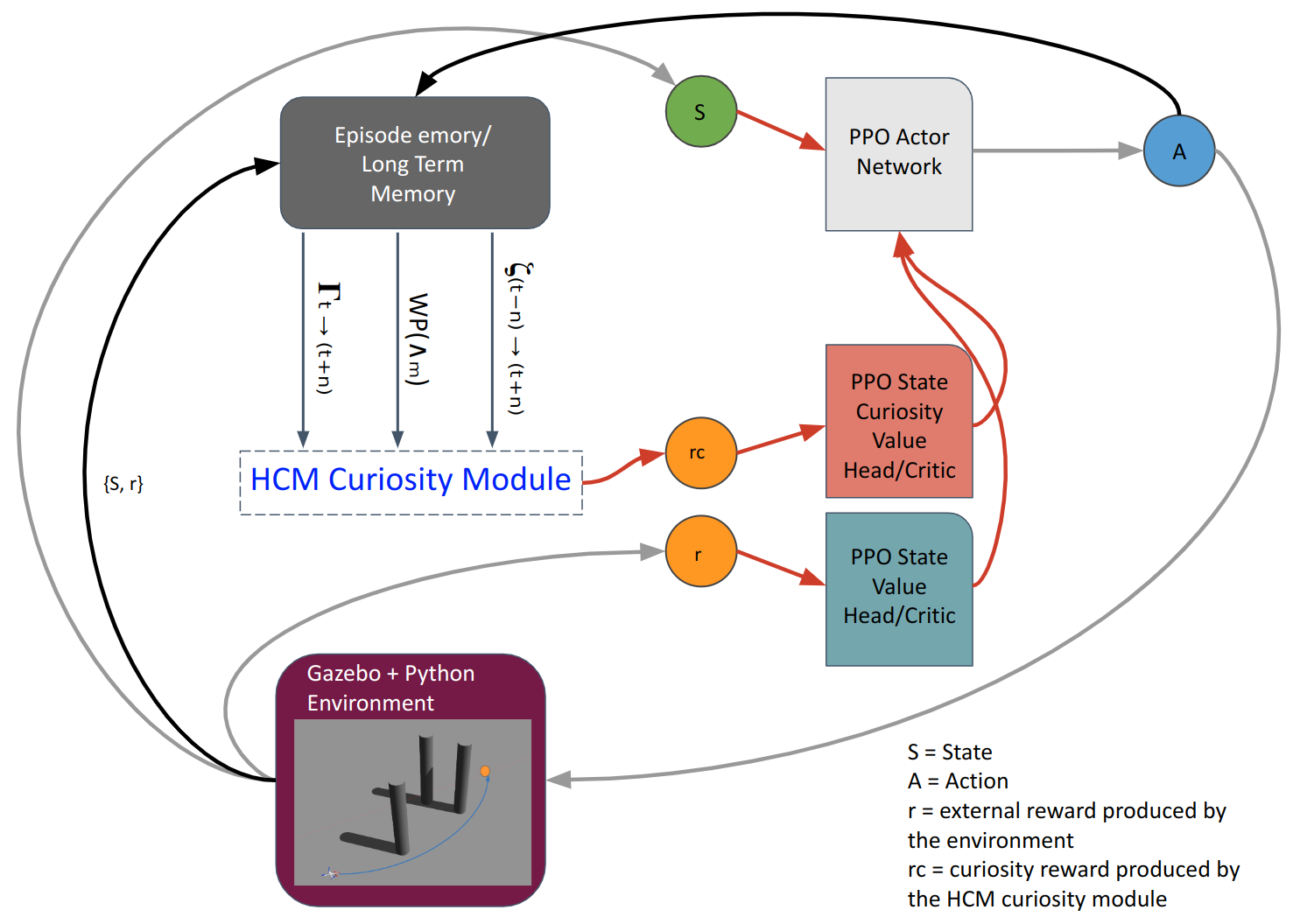}
\caption{The flow of data in our algorithm including the policy network, value networks (value heads), and curiosity module comprised of curiosity heads.}
\label{algorithm_schematic_fig}
\end{figure*}
The algorithm proposed in this paper is comprised of several parts which are illustrated in Figure \ref{algorithm_schematic_fig}.
We introduce an RL-based approach for training a curiosity-driven policy network. Curiosity is incorporated to direct the exploration of the RL method, especially in complex scenarios. Our algorithm is designed to solve the problems formulated in a Markov Decision Process (MDP) \cite{MDP} framework. The goal of the algorithm is to optimize the policy and value networks in a way that they could maximize the long-term rewards received from the environment and generated by the curiosity module. Further, an environment is designed and implemented to test the capabilities of the proposed algorithm and compare it with other approaches. In the following, different parts of the algorithm are described in detail.

\subsection{Reinforcement Learning}
The standard formulation of RL-based algorithms is used in this paper. Subsets of $s_t \in S$, $a_t \in A$, and $r_t \in R$ are defined for states, actions, and rewards. The initial starting state $D$ is defined as a set of possible initial states for the agent. Further, the following standard definitions are used throughout the paper:
\begin{eqnarray}
Q_{\pi}(s_t,a_t) &=& \mathbb{E}_{s_{t+1},a_{t+1},...} \bigg [ {\sum} ^\infty_{l=0} \gamma^l r(s_{t+1}) \bigg ] , \nonumber \\
V_{\pi}(s_t) &=& \mathbb{E}_{a_t,{s_{t+1}},...}\bigg[ {\sum}^\infty_{l=0} \gamma^l r(s_{t+1})\bigg], \nonumber \\
A_{\pi}(s,a) &=& Q_{\pi}(s,a) - V_{\pi}(s).  \nonumber
\label{rl_general_eqn}
\end{eqnarray}
where $\gamma \in (0, 1)$ is a discount factor, and $r(s_t)$ is the reward the agent receives at time $t$. 

\subsection{Proximal Policy Optimization (PPO)} 
Proximal Policy Optimization (PPO) \cite{Schulman2017-PPO} is used in this paper as the RL algorithm for optimizing the weights of the policy and value networks. 
% Considering the following conventions:
PPO is an algorithm based on and surpassing Trusted Region Policy Optimization (TRPO) \cite{Schulman2015-TRPO}. This paper briefly discusses the TRPO and PPO optimization objectives before discussing the multiple value heads used in our proposed method. 
\par
TRPO updates the policy and value networks parameters by solving the following constrained optimization problem:
\begin{eqnarray}
	& \max\limits_{\theta} \mathbb{E}_t \bigg[\frac{\pi_{\theta}(a_t \mid s_t)}{\pi_{\theta_{old}}(a_t \mid s_t)} A_t\bigg], \nonumber \\
	& \text{subject to: } \mathbb{E}_t \Big[ {KL}(\pi_{\theta_{old}}(.\mid s) \mid\mid \pi_\theta (.\mid s)) \Big] \leq \delta.
	\label{EQN TRPO 14}
\end{eqnarray}
The constraint can be incorporated in the form of a penalty weighted with a coefficient $\beta$:
\begin{eqnarray}
	& \max\limits_{\theta} 
	\hat {\mathbb{E}_t} \bigg[\frac{\pi_{\theta}(a \mid s)}{\pi_{\theta_{old}}(a \mid s)} \hat{A_t} - \beta KL[\pi_{\theta_{old}}(.\mid s), \pi_\theta (.\mid s) ] \bigg].
	\label{EQN PPO 5}
\end{eqnarray} 
The conservative policy iteration (CPI) \cite{Kakade2002} corresponds to the following surrogate objective:
\begin{eqnarray}
	& L^{CPI}{(\theta)} = 
	\hat {\mathbb{E}_t} \bigg[\frac{\pi_{\theta}(a_t \mid s_t)}{\pi_{\theta_{old}}(a_t \mid s_t)} \hat{A_t} \bigg]= 
	\hat {\mathbb{E}_t} \bigg[r_t(\theta)\hat{A_t} \bigg],
	\label{EQN PPO 6}
\end{eqnarray}
where $r_t(\theta)=\frac{\pi_{\theta}(a_t \mid s_t)}{\pi_{\theta_{old}}(a_t \mid s_t)}$ is a probability ratio comparing policy with parameters $\theta$ with the old policy with parameters $\theta_{old}$, thus, $r(\theta_{old})=1$.
\par
PPO maximizes the Equation \ref{EQN PPO 6} and penalizes changes that move $r_t(\theta)$ away from 1 using the following equation, instead of using constraint:
\begin{equation}
\begin{small}
L^{CLIP}{(\theta)} = \hat {\mathbb{E}_t} \bigg[min\bigg(r_t(\theta)\hat{A_t}, clip(r_t(\theta), 1 - \epsilon, 1 + \epsilon)\hat{A_t}\bigg) \bigg],
	\label{EQN PPO 7}
\end{small}
\end{equation}
where $\epsilon$ is a hyper-parameter, that is set equal to 0.2 in this paper.
\par
The complete loss function defined in PPO is:
\begin{equation}
\begin{small}
	L^{CLIP+VF+S}_t{(\theta)} = \hat {\mathbb{E}_t} \bigg[
	L^{CLIP}_t{(\theta)} - c_1 L^{VF}_t{(\theta)} - c_2 S[\pi_\theta](s_t)
	\bigg]
	\label{Equation PPO loss function}
\end{small}
\end{equation}
where $c_1$ and $c_2$ are hyper-parameters, $S$ is an entropy bonus similar to entropy of policy mentioned in \cite{Mnih2016a}. $L^{VF}_t$ is a squared-error loss, i.e., $L^{VF}_t = (V_\theta(s_t)-V^{targ}_t)^2$, where $V^{targ}_t=\hat{A}_t+V_{\theta_{old}}(s_t)$. To calculate $\hat{A}_t$, a truncated version of generalized advantage estimation \cite{schulman2015highdimensional} is used:
\begin{eqnarray}
	\hat{A}_t &=& \delta_t + (\gamma\lambda) \delta_{t+1} + ... + (\gamma\lambda) ^{T-t+1}\delta_{T-1}, \nonumber\\
	\delta_t &=& r_t + \gamma V_\theta(s_{t+1}) - V_\theta(s_t),
        % Old formula
	% \hat{A}_t &=& \delta_t + (\gamma\lambda) \delta_{t+1} + ... + (\gamma\lambda) ^{T-t+1}\delta_{T-1}, \nonumber\\
	% \textrm{where } \delta_t &=& r_t + \gamma V_{(s_{t+1})} - V(s_t). 
	\label{EQN PPO 11-12}
\end{eqnarray}
where $\lambda \in (0, 1)$ is an exponential discount factor.

\subsection{Multiple Value-Function Heads for PPO}
As mentioned above, our method incorporates a curiosity module which is described in Section \ref{curiosity_module_section}. We used it in our algorithm to measure the interestingness of states and direct the exploration of our algorithm. 
\par
The subtle and challenging part about a curiosity method based on neural networks is that its output changes after every $n$ episodes of training. In other words, the state value (that is, how good it is for the agent to be in a specific state) for the RL algorithm changes and is not constant such as in a standard approach where the states' values are approximated by rewards received from a constant function. The reward in such cases is the output of the reward function and is always constant for a specific pair of `state and action'. In the case of a curious algorithm, a pair of `state and action' could produce a high prediction error and be interesting at time $t$ (when the state visited for the first time by the agent), and after a couple of training epochs, the neural network output would change, resulting in a new curiosity value (which considering the training, would be less than the previous curiosity value). 
\par
Thus, in a setting with external and curiosity rewards, the RL algorithm must learn a value function that changes over time (that is, the sum of extrinsic and intrinsic values). To have a more stable learning, we use the idea of separating the parameterized value functions. We define two value heads for the PPO, the first value head called `State Extrinsic Value Head' learns the extrinsic value of the state (State Extrinsic Value) generated based on the external rewards, which is constant and comes from the environment, and the second parameterized value function called `State Intrinsic Value Head' learns the intrinsic value of the state (State Intrinsic (or Curiosity) Value), which is dynamic (that is, changes over time) and is generated based on the curiosity reward. Using this approach makes it possible to 1) stabilize the learning of the value function by using separate heads, and 2) use different learning rates for each one of the heads, thus being able to control the rate of learning the value of external reward and curiosity separately. Figure \ref{algorithm_schematic_fig} shows the flowchart of our algorithm.
\par
As a result, there is one extra loss function for the curiosity value head:
\begin{eqnarray}
L^{C}_t{(\theta)} &=& {V_{C}}_\eta(s_t)-{V_{C}}^{targ}_t, \nonumber \\
{V_{C}}^{targ}_t &=& \hat{A}_{C_t} + V_{C_{\eta_{old}}}(s_t).
	\label{Equation PPO Curiosity Part loss function}
\end{eqnarray}
Further, the $\delta_t$ estimation mentioned in Equation \ref{EQN PPO 11-12} would change as the following:
\begin{eqnarray}
\delta_t &=& (r_{ext} + r_{int}) + \gamma \bigg((V(s_{t+1}) - V(s_t)\bigg) \nonumber \\
& & + \bigg({V_{C}}(t+1)-{V_{C}}(t)\bigg).
\label{EQN PPO Advantage Update}
\end{eqnarray}
$r_{ext}$ and $r_{int}$ are rewards generated by the environment and curiosity module and are described in Section \ref{reward_section}.

\subsection{Training a Quad-copter}
We designed and implemented an environment (explained thoroughly in Section \ref{environment_section}) to evaluate the performance of the proposed method and to compare it with that of other methods. In the case of a quad-copter control using reinforcement learning, the agent should generate 100 actions per second (100 Hz) and collect the state, action, and reward in each time-step. We execute the training at the end of each episode of 16,394 steps, mainly because it is very stable, but the algorithm can be trained in every 4,096 steps or 8,192 steps, where the training would be less stable.

\subsection{Policy and Value Networks}
The policy network is a neural network with two hidden layers, each formed by 256 neurons. The input to the network (the agent state or perception of the world) is formed by the odometry data (including both linear and angular accelerations), the previous motor speeds (that can be considered as the proprioception of quad-copters' motors state), the distances of the obstacles to the agent, and the distance of the agent to the goal. Distances are calculated based on the X and Y coordinates. For each obstacle the distance comprises of the distance in X coordinate, the distance in Y coordinate, and Euclidean distance based on the previous two measures, while the distance to the goal is comprised of only the Euclidean distance based on X and Y. The odometry part of the state is introduced as input to the network, while the rest of the state is concatenated to the output of the first layer, which is subsequently introduced to the second layer of the network. The output of the policy network is a 4-dimensional vector that is used to generate the action. An identical structure is used for the state value network and the curiosity value network, except for the output, which is the state value of a particular observation. All networks are optimized using the Proximal Policy Optimization (PPO) algorithm.
\begin{figure}[htbp]
\centering
\includegraphics[width=0.3\textwidth]{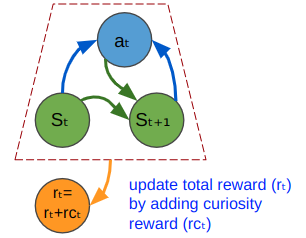}
\caption{The dynamics between state, action, and reward in the ICM \cite{DBLP:journals/corr/PathakAED17}. The method is discussed briefly at the beginning of Section \ref{ICM_section}.}
\label{curiosity_original_method_fig}
 % where $\phi$ is a function for extracting the state features, and $X$ is a scaling factor. 
\end{figure}

\subsection{Intrinsic Curiosity Module}
\label{ICM_section}
We used the idea of curiosity based on the prediction error similar to the method in \cite{DBLP:journals/corr/PathakAED17}. In the original Intrinsic Curiosity Module (ICM) method, two models are defined for calculating curiosity. The first model, called inverse dynamics, is used to learn to predict the action that the agent took to move from $S_t$ to $S_{t+1}$, i.e.,:
\begin{eqnarray}
&\hat{a}=g(s_t, s_{t+1};\theta_I),
\end{eqnarray}
and a loss function is used for reducing the error between the predicted action and the actual action:
\begin{eqnarray}
\min_{\theta_I}L_I(\hat{a}_t, a_t).
\end{eqnarray}
Further, there is a forward dynamics model that is responsible for the prediction of the state feature $\phi(s_{t+1})$ based on the state $\phi(s_{t})$ and action $a_t$:
\begin{eqnarray}
&\hat{\phi}(s_{t+1})=f(\phi(s_t), a_t; \theta_F)
\end{eqnarray}
with its corresponding loss function being:
\begin{eqnarray}
&L_F(\phi(s_t), \hat{\phi}(s_{t+1}))=\frac{1}{2}\left \| \hat{\phi}(s_{t+1}) - \phi(s_{t+1}) \right \|^{2}.
\end{eqnarray}
The intrinsic (curiosity) reward in each time-step $r_{t}^{i}$ is then defined to be the error of prediction between the actual feature sate $\phi(s_{t+1})$ and the predicted feature state $\hat{\phi}(s_{t+1})$, i.e.,:
\begin{eqnarray}
&r_{t}^{i}=\frac{\eta}{2}\left \| \hat{\phi}(s_{t+1}) - \phi(s_{t+1}) \right \|.
\end{eqnarray}
The overall loss function for the curiosity module is defined to be:
\begin{eqnarray}
\min_{\theta_I,\theta_F}\bigg [ (1-\beta)L_I + \beta L_F \bigg].
\label{Equation curiosity original method}
\end{eqnarray}
Figure \ref{curiosity_original_method_fig} shows a schematic of the dynamics of the described approach. 
\par
Overall, while the above-described method is a capable approach useful in some problems, it cannot help our algorithm learn to maximize the reward and, as shown in Section \ref{evaluation_section}, it fails to effectively train the agent. Therefore, we propose a new curiosity approach, explained in the next section.

\subsection{High-level Curiosity Module}
\label{curiosity_module_section}
The approach this paper proposes for calculating curiosity is based on the prediction error with the following modifications:
\begin{itemize}
    \item A new state space for the curiosity module that uses the segment of states instead of a normal state.
    \item A pre-processing function that transforms the segment of actions before using it by the curiosity module.
    \item In addition to predicting the curiosity using the dynamics in the state and action spaces, our method uses the dynamics between the state, action, and external rewards for predicting the curiosity.
    \item Instead of using a single network, our algorithm incorporates an ensemble of curiosity networks, as shown in Figure \ref{curiosity_module_fig}.
    \item Finally, instead of calculating a curiosity reward for a single step, our method calculates and updates a trajectory of the steps using a decay factor.
\end{itemize}
We call it High-level Curiosity Module (HCM) because it learns based on the trajectory of the agent's interaction with the environment. As a result, it is a high-level approach compared to learning from a single interaction. Further, as Curiosity is an intrinsic value by nature, we do not use the intrinsic name in the title of the algorithm.

\subsection{Curiosity in States-Actions-States Space}
Considering a low-level control problem where the agent receives sensor data 100 times per second, the changes between the states would be minor, generating a proper curiosity reward would be challenging. One way to address this issue is to modify the state, action, and reward space for the curiosity module. In order to make a curious agent that can converge to an optimal solution, we create a segment of states called $\zeta$ where it can select a set of states in backward pass or forward pass, considering the current time $T=t$. We create a segment of the trajectory of the agent with length $n$ steps starting back in time and ending at the current time-step, i.e., $\zeta_{(t-n)\rightarrow t}=\{s_{t-n},s_{t-(n-1)}, ..., s_{t-0}\}$. We also create a second segment of the trajectory of the agent starting at the current time-step and ending $n$ steps into the future, i.e., $\zeta_{t \rightarrow (t+n)}=\{s_t,s_{t+1}, ..., s_{t+n}\}$. 
Intuitively, it is necessary to use a segment of actions for the transition between $\zeta_{(t-n)\rightarrow t}$ to $\zeta_{t \rightarrow (t+n)}$, which we call it $\Lambda$. While all the actions for moving from $\zeta_{(t-n)\rightarrow t}$ to $\zeta_{t \rightarrow (t+n)}$ are effective, the actions that are more near to the connection of the two segments (or trajectories) are more important. Thus, considering only half of the actions from each segment would be enough. So, $\Lambda_m = \{a_{t-m}, ..., a_{t+m}\}$, i.e., it comprises of the set of actions starting from time-step $T=t-m$ to $T=t+m$, where $m=n/2$. However, training a neural network to predict $\Lambda_m$ based on $\zeta_{(t-n)\rightarrow t}$ and $\zeta_{t \rightarrow (t+n)}$ (that is, $\Lambda_m = g_{SS}(\zeta_{(t-n)\rightarrow t},\zeta_{t \rightarrow (t+n)}; \theta_{I_{SS}})$ where $\theta_{I_{SS}}$ is the set of the $g_{SS}$ model parameters) is not trivial and does not produce good results, mainly because of the size of $\Lambda_m$. 
% Old part
%Further, it is necessary to use a segment of actions for the transition between $\zeta_{(t-n)\rightarrow t}$ to $\zeta_{t \rightarrow (t+n)}$, and we call it $\Lambda_m = \{a_{t-m}, ..., a_{t+m}\}$ that comprises of a set of actions starting from time-step $T=t-m$ to $T=t+m$, where $m=n/2$. However, training a neural network to predict $\Lambda_m$ based on $\zeta_{(t-n)\rightarrow t}$ and $\zeta_{t \rightarrow (t+n)}$ is not trivial and does not produce good results. 
A better solution is to first convert the $\Lambda_m$ to a low-dimensional vector that characterizes the transition. To do this, we define a function called $F_{WP}$. 
\par
% Added part
One general approach would be to define $F_{WP}$ as a convolutional neural network or variational auto-encoder that extracts the features or the latent space of the Actions segment and consider it as the low-dimensional vector that characterizes the transition.
However, considering that we have access to positions and attitudes in our problem, an easier approach is to directly calculate the waypoints that indicate the transition. A waypoint could be considered as a `position' or a combination of `position and attitude' transition. Here, we consider it as a `position' transition because our goal is to reduce the dimensionality of the Actions segment. Thus, instead of using the action space to characterize the transition, we consider the change in position space as the transition. Moreover, to capture the transition adequately well, more than one waypoint is needed. We use three waypoints in order to have a measure from the beginning, middle, and end of the transition. As a result, the $F_{WP}$ input is $\Lambda_m$, a subset of $\{\zeta_{(t-m)\rightarrow t}, \zeta_{t \rightarrow (t+m)}\}$ comprised of only position data where $m=n$. $m$ could be smaller or larger than $n$ in general approach, such as $m=n/2$. The output of $F_{WP}$ is $\{wp_1,wp_2,wp_3\}$.

Our parameterized inverse dynamics function called $g_{SS}$ predicts the output of $F_{WP}$ and is defined as follows:
\begin{eqnarray}
&\hat{F}_{WP}(\Lambda_m) = g_{SS}(\zeta_{(t-n)\rightarrow t},\zeta_{t \rightarrow (t+n)}; \theta_{I_{SS}}),
\end{eqnarray}
where $\theta_{I_{SS}}$ is the set of the $g_{SS}$ model parameters, and the loss function becomes:
\begin{eqnarray}
\min_{\theta_{I_{SS}}}L_{I_{SS}}(\hat{F}_{WP}(\Lambda_m), F_{WP}(\Lambda_m)).
\end{eqnarray}
Further, the forward dynamics are defined as follows:
\begin{eqnarray}
&\hat{\phi}(\zeta_{t \rightarrow (t+n)})=f_{SS}\bigg(\phi(\zeta_{(t-n)\rightarrow t}), F_{WP}(\Lambda_m); \theta_{F_{SS}}\bigg),
\end{eqnarray}
where $\phi$ is the function to extract the features, and $\theta_{F_{SS}}$ is a set of the $f_{SS}$ model parameters. We then minimize the following loss function:
\begin{eqnarray}
& L_{F_{SS}}(\phi(\zeta_{t \rightarrow (t+n)}), \hat{\phi}(\zeta_{t \rightarrow (t+n)}))= \nonumber \\
& \frac{1}{2}\left \| \hat{\phi}(\zeta_{t \rightarrow (t+n)}) - \phi(\zeta_{t \rightarrow (t+n)}) \right \|^{2}.
\end{eqnarray}
The curiosity reward generated by this head is calculated in the following way:
\begin{eqnarray}
r_{C_{SS}} = \bigg [ (1-\beta)L_{I_{SS}} + \beta L_{F_{SS}} \bigg].
\label{Equation curiosity module part A}
\end{eqnarray}
Finally, the total loss function is:
\begin{eqnarray}
\min_{\theta_{I_{SS}},\theta_{F_{SS}}}\bigg [ r_{C_{SS}} \bigg].
\label{Equation curiosity module loss part A}
\end{eqnarray}

\subsection{Curiosity in States-Actions-Rewards Space}
Considering that an immediate reward exists in our problem setting, generating the curiosity reward would be more complex because the policy constantly receives the reward from the environment, making the problem of generating a proper curiosity reward more challenging. In order to make a curious agent that can converge to the optimal solution, we define a new pair of forward and inverse dynamics models where instead of predicting the next state, they would work and predict the segment of external reward. In a nutshell, we use the $\zeta_{(t-n)\rightarrow t}$ and $F_{WP}$ as they described in the previous section. Further, we define a segment of external rewards and show it by $\Gamma_{t \rightarrow (t+n)}$. Also, we define a new set of functions for inverse and forward dynamics:
\begin{eqnarray}
&\hat{F}_{WP}(\Lambda_m) = g_{SR}(\zeta_{(t-n)\rightarrow t},\Gamma_{t \rightarrow (t+n)}; \theta_{I_{SR}}),
\end{eqnarray}
where $\theta_{I_{SR}}$ is set of the $g_{SR}$ model parameters, and the loss function
\begin{eqnarray}
\min_{\theta_{I_{SR}}}L_{I_{SR}}(\hat{F}_{WP}(\Lambda_m), F_{WP}(\Lambda_m)).
\end{eqnarray}
Further, the forward dynamics are defined as follows:
\begin{eqnarray}
&\hat{\phi}(\Gamma_{t \rightarrow (t+n)})=f_{SR}(\phi(\zeta_{(t-n)\rightarrow t}), F_{WP}(\Lambda_m); \theta_{F_{SR}}),
\end{eqnarray}
where $\phi$ is the function to extract the features, and $\theta_{F_{SR}}$ is a set of the $f_{SR}$ model parameters. We then minimize the following loss function:
\begin{eqnarray}
& L_{F_{SS}}(\phi(\Gamma_{t \rightarrow (t+n)}), \hat{\phi}(\Gamma_{t \rightarrow (t+n)}))= \nonumber \\
& \frac{1}{2}\left \| \hat{\phi}(\Gamma_{t \rightarrow (t+n)}) - \phi(\Gamma_{t \rightarrow (t+n)}) \right \|^{2}.
\end{eqnarray}
The curiosity produced by this head is:
\begin{eqnarray}
r_{C_{SR}} = \bigg [ (1-\beta)L_{I_{SR}} + \beta L_{F_{SR}} \bigg],
\label{Equation curiosity module part B2}
\end{eqnarray}
and the total loss function is:
\begin{eqnarray}
\min_{\theta_{I_{SR}},\theta_{F_{SR}}}\bigg [ r_{C_{SR}}  \bigg].
\label{Equation curiosity module part B1}
\end{eqnarray}

\subsection{Ensemble of Curiosity Sub-modules}
\begin{figure}[htbp]
\includegraphics[width=0.5\textwidth]{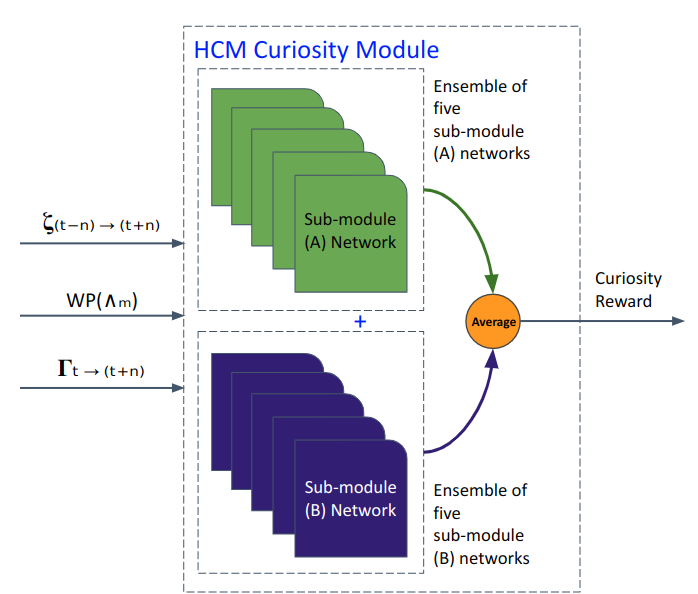}
\caption{Overall architecture of our curiosity module where n Module A and n Module B heads are used to create the curiosity module.}
\label{curiosity_module_fig}
\end{figure}

In the previous sections, two curiosity sub-modules were defined, namely 1) curiosity error based on prediction error in $(\zeta, WP(\Lambda_m), \zeta)$ transition space (called curiosity module part a), and 2) curiosity error based on the error of prediction in $(\zeta, WP(\Lambda_m), \Gamma)$ transition space (called curiosity module part b). To have a less biased curiosity module, we define an ensemble of curiosity similar to \cite{Balaji2017} in terms of using multiple networks to measure the prediction error. In our setting, our final curiosity module comprises of five curiosity sub-modules type `a' and five curiosity sub-module type `b', as illustrated in Figure \ref{curiosity_module_fig}. The output of the curiosity module then is the average of all networks' outputs, and the final formula for calculating the curiosity reward consisting of $n$ ensemble of networks (for each separate sub-module) is:
\begin{equation}
r_{curiosity}=\alpha_{curiosity} * \bigg(\frac{1}{2n}\Sigma(r_{C_{SS_n}} + r_{C_{SR_n}})\bigg),
\end{equation}
where $\alpha_{curiosity}$ is a coefficient for curiosity reward.
\begin{figure}[htbp]
\includegraphics[width=0.5\textwidth]{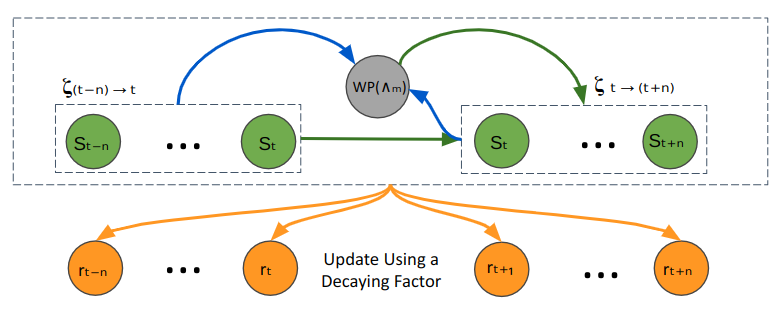}
\caption{Part one of our proposed curiosity method where the prediction error is based on the state segments and action segments.}
\label{curiosity_module_a_fig}
\end{figure}
\begin{figure}[htbp]
\includegraphics[width=0.5\textwidth]{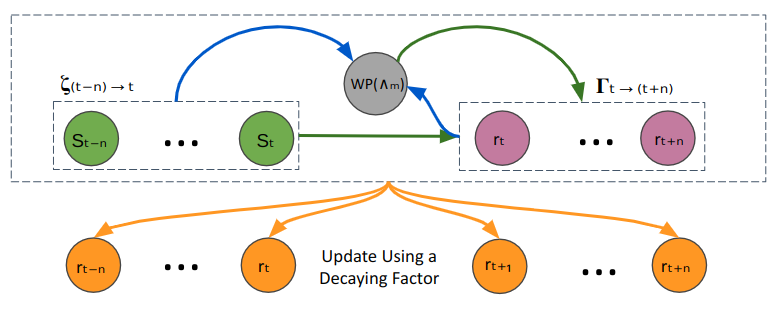}
\caption{Part two of our proposed curiosity method where the prediction error is based on the state segments, action segments, and reward segments.}
\label{curiosity_module_b_fig}
\end{figure}
\begin{figure*}[th]
\includegraphics[width=0.99\textwidth]{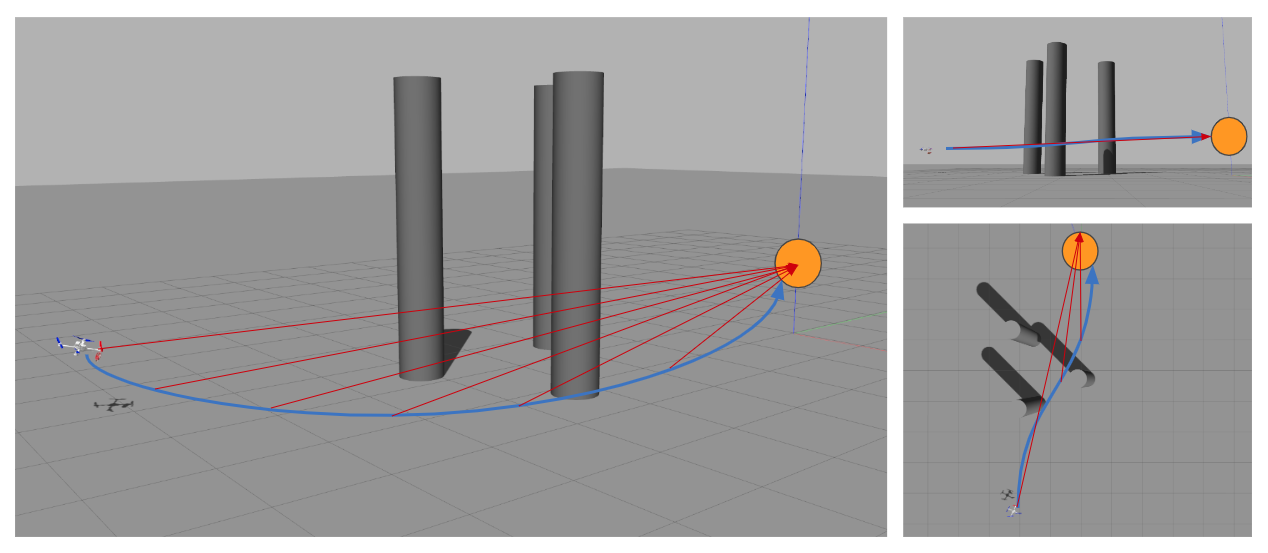}
\caption{Visualization of our environment (3D view (left), side view (top right), and top view (bottom right)), which is comprised of a quad-copter and three obstacles initialized in a random position between the drone and the desire position. The goal is to 1) pass the obstacles and reach a goal destination known by the algorithm by a position and attitude pair (the blue line draws a sample desired path), and 2) control the Yaw towards the desired position (in this case, the coordinate frame origin), the red lines visualize the desired Yaw directions.}
\label{new_environment}
\end{figure*}
\begin{figure}[htbp]
\centering
\includegraphics[width=0.4\textwidth]{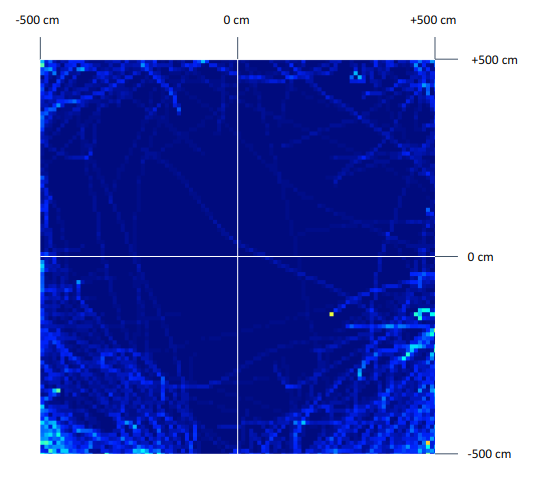}
\caption{The trajectories traversed by the agent in the environment from a top-view point of view (i.e., in the XY plane) after 50 episodes.}
\label{curiosity_visualization_sample_fig}
\end{figure}

\subsection{Updating a Trajectory of State Curiosity Values}
Considering that the curiosity reward generated by our method is generated based on $(\zeta, WP(\Lambda), \Gamma)$, a segment of states, actions, and rewards, we convert the generated curiosity reward to a trajectory of curiosity rewards and use it to update a trajectory of State Curiosity Values as follows:
\begin{eqnarray}
r_{c_{t \pm x}}=r_{c_{t \pm x}} + \kappa^x(r_{curiosity}),
\end{eqnarray}
where $x=0,1,..,n$ and $\kappa \epsilon (0, 1)$ is a decay factor.

\subsection{Visualizing the Effect of Curiosity}
\label{curiosity_visualization_section}
One way to visualize the effect of curiosity is to trace and visualize the trajectories traversed by the robot/agent, which is the method we used and described here. We first consider the XY plane of the agent position and trace its movement in that plane. In order to trace its movement, we divide the XY plane into X columns and Y rows and define a visitation value for each cell. Further, each time the agent passes through a row and column, we add the visitation value of that cell by +1. Finally, we normalize the matrix that collects this information and visualize it as an image. A sample image that shows the agent's movement in the environment is shown in Figure \ref{curiosity_visualization_sample_fig}.

\subsection{Environment}
\label{environment_section}
The environment is an essential part of an RL-based approach considering it generates the new state $s_t$ and external reward $r_t$ by executing the action $a_t$. The environment used in this work is based on the Gazebo simulator \cite{Gazebo} and the RotorS package \cite{Furrer2016}, which is used to simulate a model of Ascending Technology Hummingbird. The environment is comprised of a quad-copter and three obstacles. To make the environment realistic and reduce the gap between the simulation and real world our environment adds noise to the motors and simulate the damping effect. The primary goal for the quad-copter agent is to learn to fly and control itself from any random initial position and attitude. The secondary goal is explained using the following items:
\begin{itemize}
    \item Pass the obstacles and fly towards the desired position and attitude.
    \item Control the yaw direction toward the desired position in terms of X and Y.
\end{itemize}
Besides that, a Python script is used to communicate with the Gazebo simulator to retrieve the new state, generate the extrinsic rewards after passing the action generated by the policy network, detect the terminal states and reset the quad-copter position, and gather all the statistical information related to the quad-copter, the extra goals, and errors. The communication between the Python script and the Gazebo simulator is handled by using Robot Operating System (ROS) \cite{ROS} to enhance the mobility of the code. 
\subsection{Reward}
This work defines three types of rewards for training our RL-based policy:
\begin{itemize}
    \item An immediate extrinsic reward is generated by the environment in each time step; this reward is a measure of how near the agent is to the desired position and attitude and is defined using the following formula:
    \begin{eqnarray}
    % r_{flight} &= -\alpha_p(|x_d-x_c| + |x_d-x_c| + |x_d-x_c|)  \nonumber \\
    & r_{flight} = \nonumber \\
    & \alpha_p\bigg((1-\frac{|x_d-x_c|}{x_{max}}) + (1-\frac{|y_d-y_c|}{y_{max}}) + (1-\frac{|z_d-z_c|}{z_{max}})\bigg)  \nonumber \\
    &  - \alpha_a \bigg(|\theta_d-\theta_c|+|\phi_d-\phi_c|+|\psi_d-\psi_c| \bigg),\hspace{5pt}
    \end{eqnarray}
    where $(x, y, z)$ denotes the position coordinates and $(\theta, \phi, \psi)$ are the attitude of the agent. The index $d$ indicates the desired position and attitude, which are $x_d=0m$, $y_d=0m$, $z_d=1.5m$, $\theta_d=0$, $\phi_d=0$, and $\psi_d$ is equal to the Yaw that points towards the origin of the world coordinate system. The index $c$ indicates the current position and attitude. $\alpha_p$ and $\alpha_a$ are coefficients for error in position and error in attitude, respectively. Finally, we use a shaping reward technique to motivate agent activity in the desired areas by defining a threshold for x, y, and z axes. If the agent distance in a specific axis gets more than a predefined threshold, the position reward for that axis would be $-1.0$. For example, if $|x_d-x_c|>\frac{x_{max}}{2}$ then we consider $-1.0$ instead.
    \item Two auxiliary immediate extrinsic rewards that are generated by the environment. The first one is related to the control of the yaw toward the goal direction, as shown in the following: 
    \begin{eqnarray}
    & r_{yaw}= |\psi_d-\psi_c|,
    \end{eqnarray}
    where $\psi$ is the Yaw of the attitude of the agent, the index $d$ indicates the desired Yaw value, and the index $c$ the current Yaw value.
    \\
    The second one is related to the velocity of the agent as shown in the following: 
    \begin{eqnarray}
    & r_{vel}= \alpha_{\nu}*||\nu||+\alpha_{\omega}*||\omega||,
    \end{eqnarray}    
    where $\nu$ and $\omega$ are the linear and the angular velocity of the quad-copter, respectively, and $\alpha_{\nu}$ and $\alpha_{\omega}$ are coefficients.
    \item An intrinsic immediate reward that is generated by the curiosity module. This reward represents the surprise and motivates the agent's exploration and is fully explained in Section \ref{curiosity_module_section}.
\end{itemize}
Finally, the final reward is calculated in two parts, considering the two Value Heads defined for the PPO (i.e., State Value Head and State Curiosity Value Head), the internal reward explained in Section \ref{curiosity_module_section}, and the external reward is formulated as shown in the following:
\begin{eqnarray}
  {r_{ext}}=\begin{cases}
      (r_{flight} * \alpha_{flight}) + 
        (r_{yaw} * \alpha_{yaw}) +
        r_{vel},\\
    -10 \text{,  if quad-copter hits obstacle or crashes}.
  \end{cases}
	\label{Equation both rewards definition}
\end{eqnarray}
\label{reward_section}
% \begin{figure}[th]
% \includegraphics[scale=0.25]{sections/images/rewards_ptl_idealistic.png}
% \caption{This figure shows that point-to-line negative reward is reduced through episodes (an idealistic environment).}
% \label{point_to_line_image}
% \end{figure}
% \begin{figure*}[th]
% \includegraphics[scale=0.24]{sections/images/rewards_all_realistic.png}
% \caption{The left part of this figure shows reward maximization (in an realistic environment), and the right side shows curiosity rewards.}
% \label{rewards_with_noise}
% \end{figure*}
\section{Experiments}
\label{experiments_section}
In order to illustrate the performance of the proposed algorithm we compared it with other powerful algorithms which serve as the baselines:
\begin{itemize}
  \item PPO is used as the baseline on-policy algorithm for testing and performance comparison in our tests. 
  \item SAC is used as an off-policy algorithm with memory replay to compare its performance with the other on-policy algorithms mentioned in this section.
  \item PPO+ICM is the PPO algorithm combined with ICM \cite{DBLP:journals/corr/PathakAED17} module (for the curiosity reward). 
  \item PPO+HCM is the PPO algorithm combined with our proposed curiosity approach (i.e., HCM).
\end{itemize}
The PPO algorithm we used here as the baseline is a highly tuned PPO algorithm against our fly environment with exploration noise generated by a Gaussian distribution with a mean of 0 and standard deviation of 1.0. The default parameters used for the SAC algorithm (the stable baseline version). 
We tested the performance of the competing algorithms by running our scenario multiple times. Each algorithm was run six times, and the min, max and average results are calculated. Our code can be found in the corresponding GitHub repository\footnote{\url{https://github.com/a-ramezani/CDRL-L2FC_u_HCM}}.

% \begin{figure}[h]
% \includegraphics[scale=0.33]{sections/images/att-err.png}
% \label{attitude_error_image}
% \caption{This figure shows the attitude error.}
% \end{figure}

%\section{Evaluation and Results}
\label{evaluation_section}
\begin{figure*}[]
\includegraphics[width=0.99\textwidth]{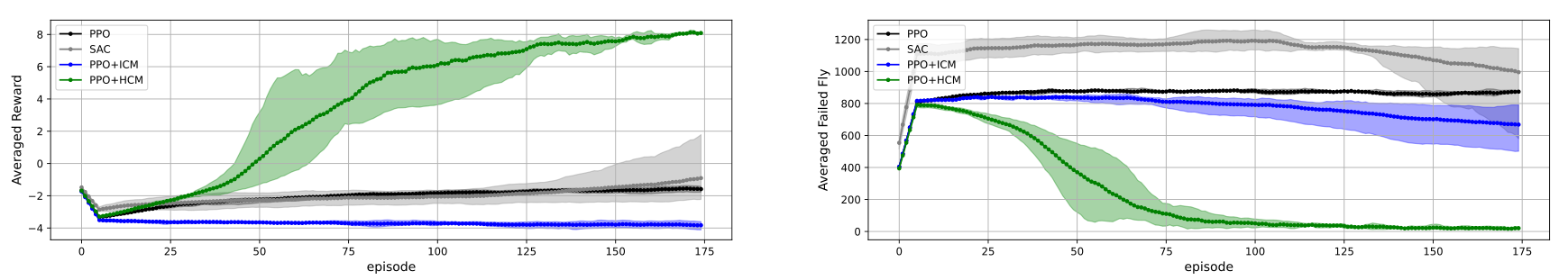}
\caption{Capability of each algorithm in maximizing the reward through time (left). Capability of each algorithm to reduce the Failed Fly (crashes) measure (right).}
\label{averaged_reward_averaged_failed_fly_fig}
\end{figure*}
\begin{figure*}[th]
\includegraphics[width=0.99\textwidth]{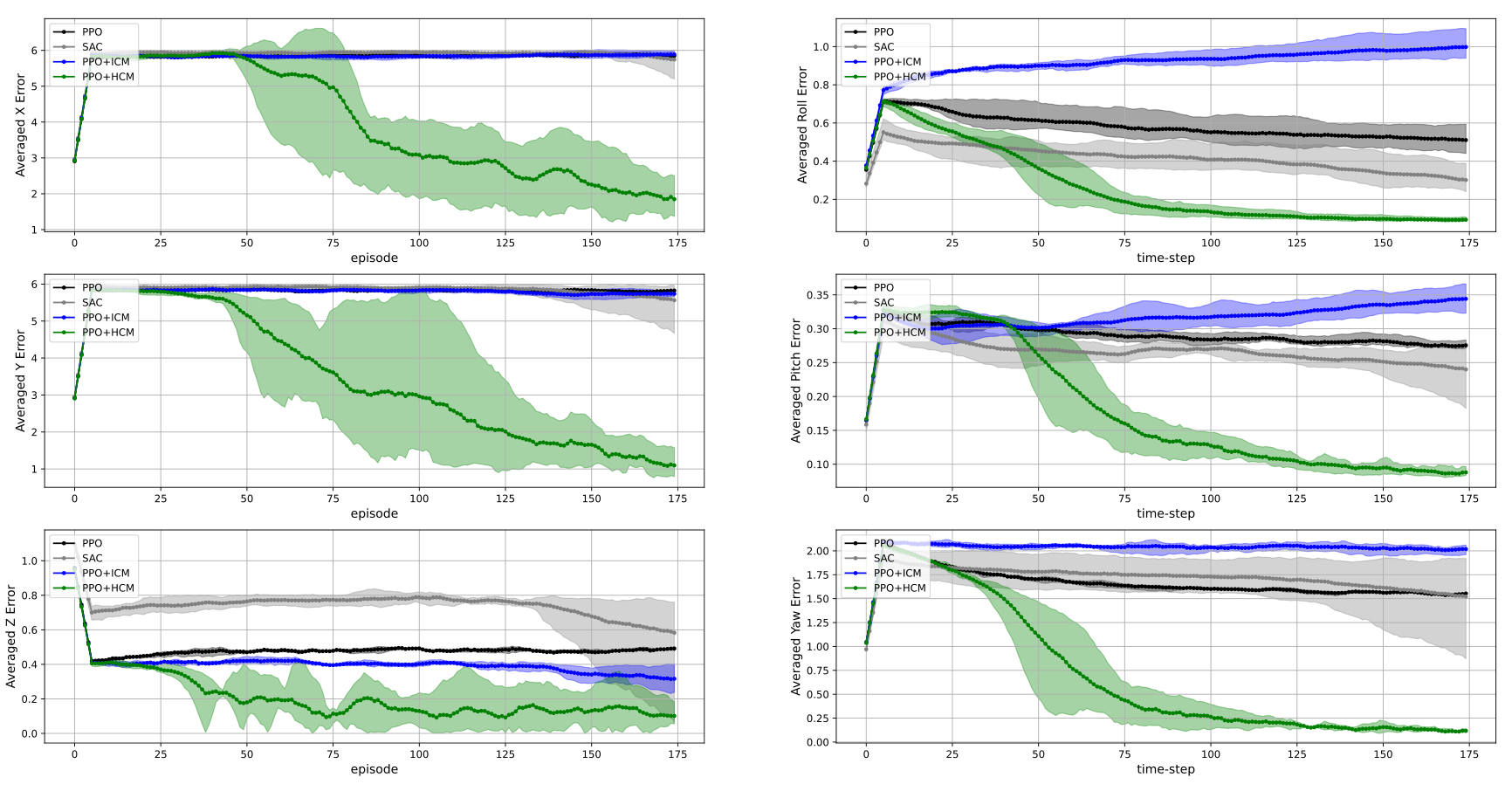}
\caption{Position and attitude errors: position errors for X, Y, and Z (left), and attitude error in Roll, Pitch, and Yaw (right).}
\label{averaged_errors_fig}
\end{figure*}
\begin{figure*}[]
\includegraphics[width=0.99\textwidth]{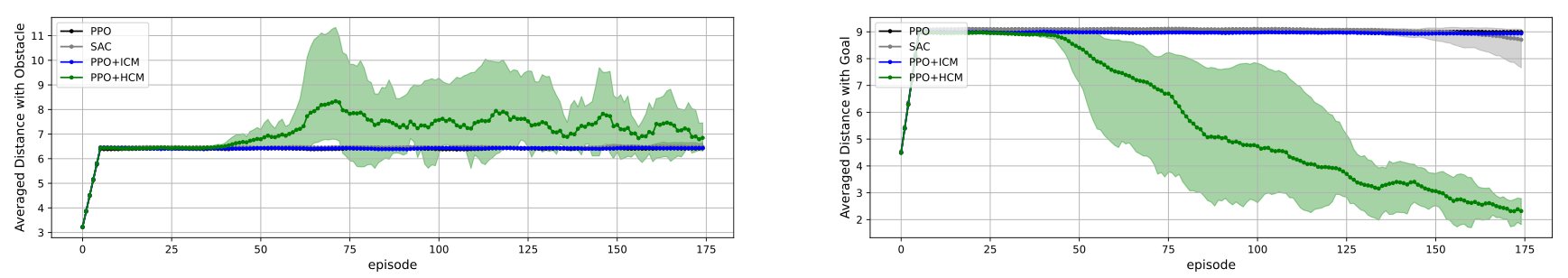}
\caption{Average distance between the quad-copter and the obstacles (left), and average distance between the quad-copter and the goal position (right).}
\label{averaged_distances_goal_obstacle_fig}
\end{figure*}
\begin{figure*}[]
\centering
\includegraphics[width=0.99\textwidth]{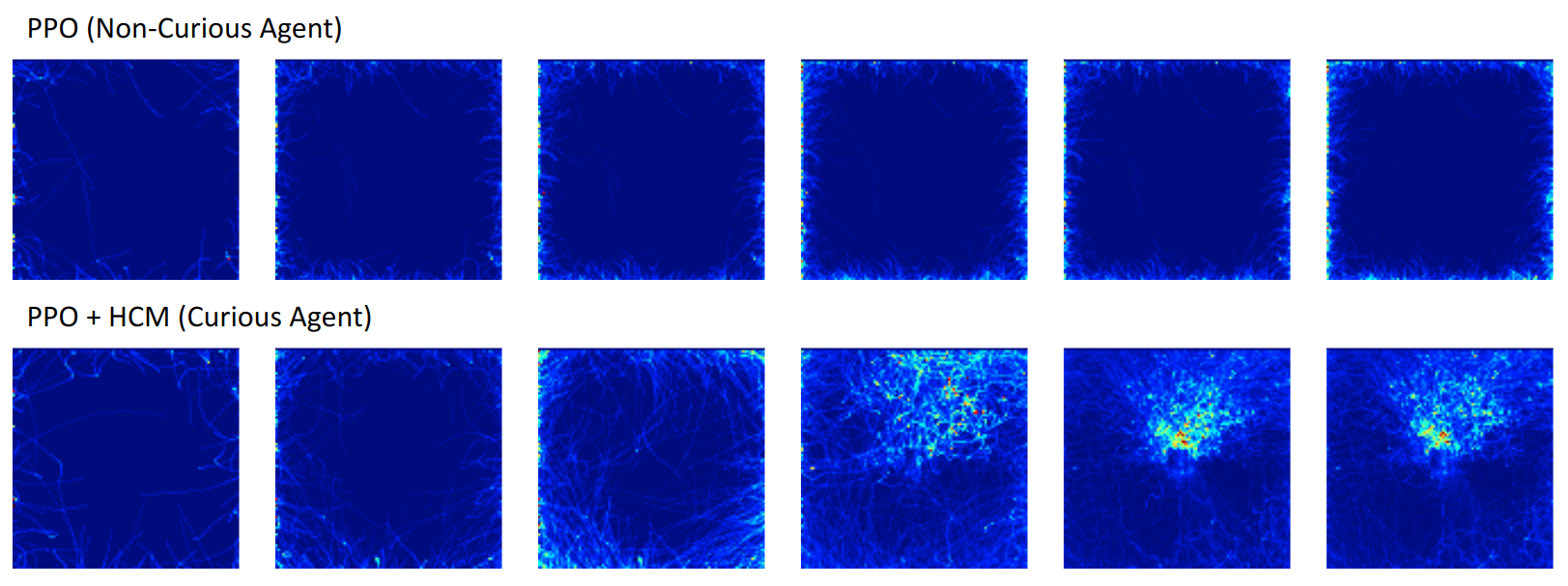}
\caption{Visualization of the effect of curiosity in the agent's trajectories (in XY plane, i.e., top-view). Starting from left to right of the figure, by increasing the number of episodes, the curious agent would have more activities toward the center of the image or toward the desired position, but the non-curious agent avoids moving toward the center mainly because it avoids hitting the obstacles. This figure focuses only on the areas important for exploration.}
\label{curiosity_visualization_fig}
\end{figure*}
\begin{figure*}[]
\centering
\includegraphics[width=0.99\textwidth]{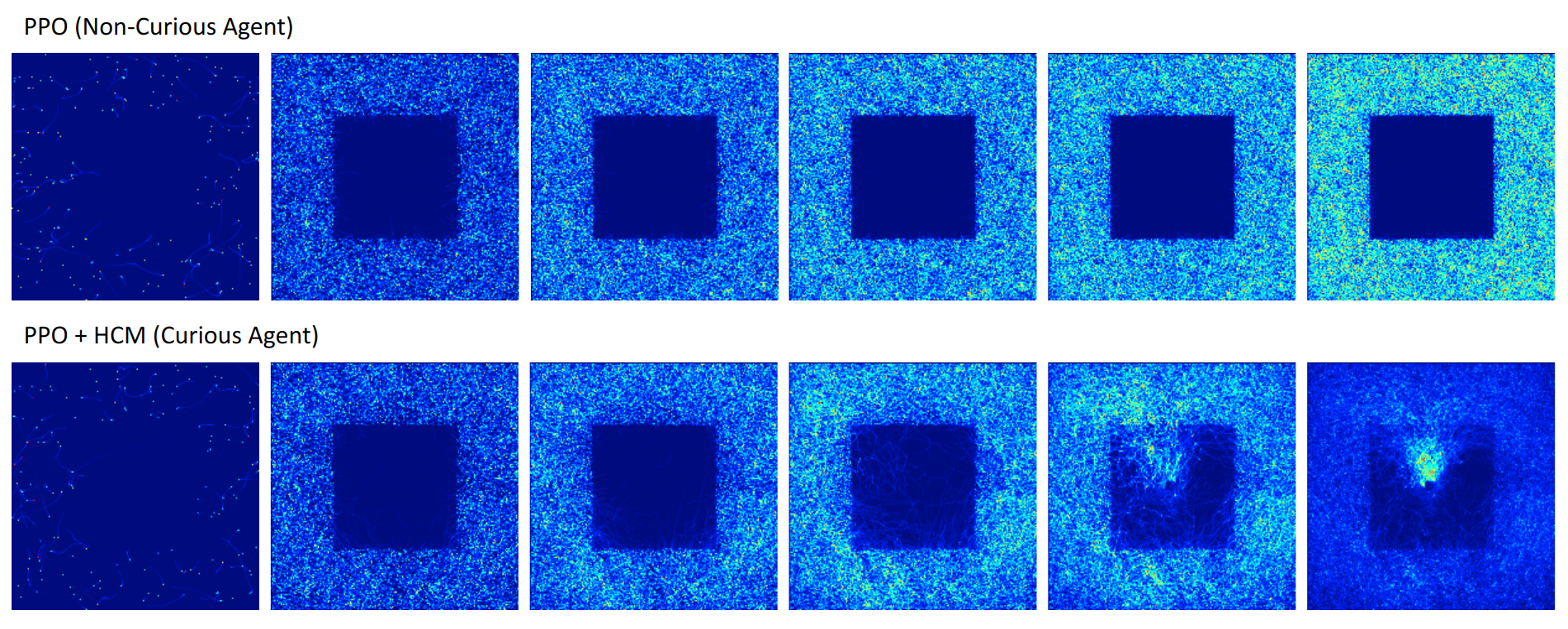}
\caption{Visualization of the effect of curiosity in the agent's trajectories (in XY plane, i.e., top-view). Starting from left to right of the figure, by increasing the number of episodes, the curious agent would have more activities toward the center of the image or toward the desired position, but the non-curious agent avoids moving toward the center mainly because it avoids hitting the obstacles. This figure visualizes on all the areas traversed by the agent.}
\label{curiosity_visualization_fig_inc}
\end{figure*}

\begin{figure*}[]
\centering
\includegraphics[width=0.99\textwidth]{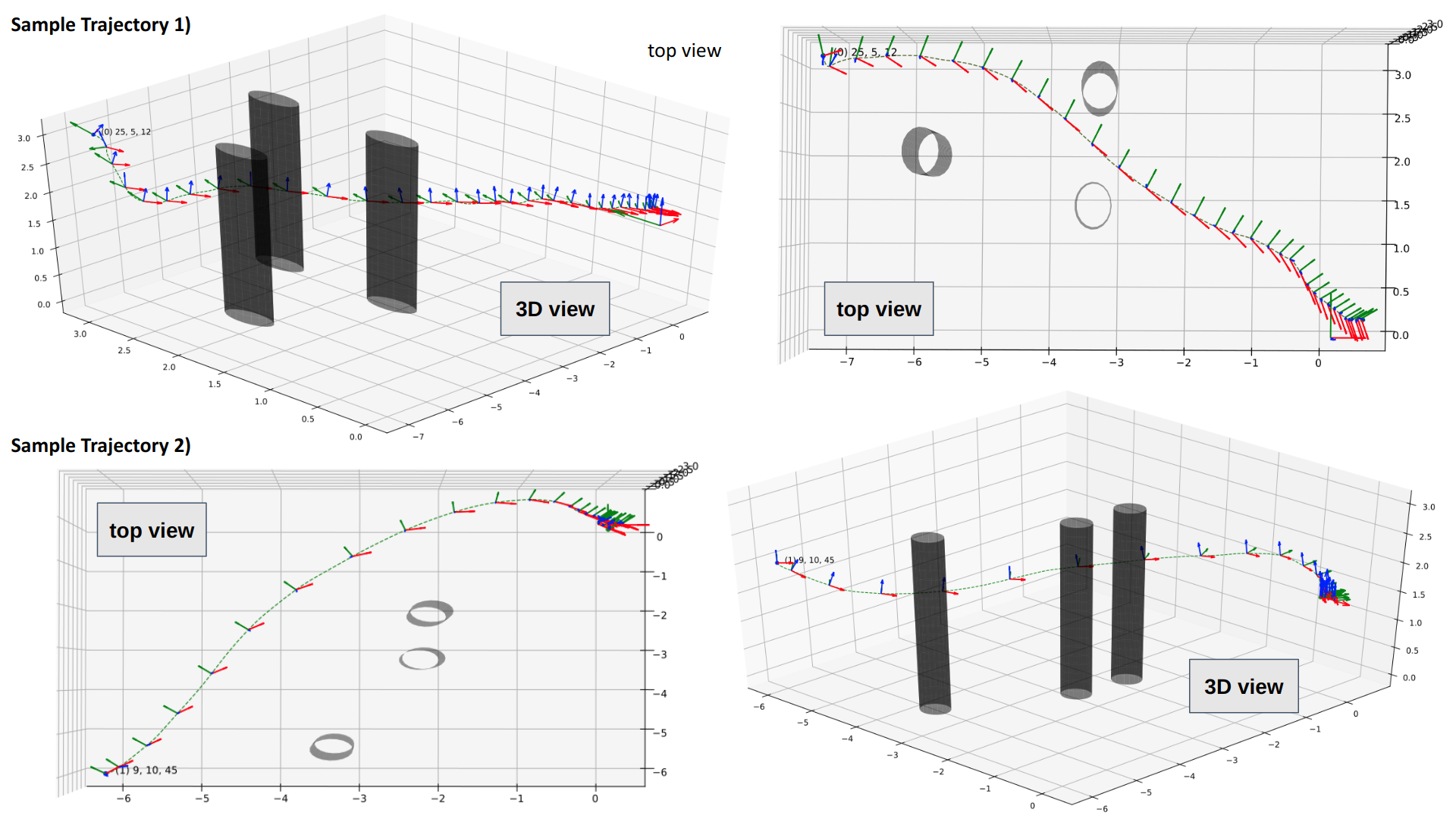}
\caption{Visualization of two sample trajectories of our optimal policy (the top view and 3D view of each trajectory are shown per row). The quad-copter is initialized in random position and attitude, and obstacles are initiated in random positions.}
\label{optimal_policy_trajectories_fig}
\end{figure*}

\subsection{Reward Maximization and Quad-copter Low-level Control}
Reward maximization is the main goal in reinforcement learning-based algorithms. Figure \ref{averaged_reward_averaged_failed_fly_fig} illustrates the performance of the competing algorithms. Looking at the left side of the figure, one can observe that only PPO+HCM (i.e., our proposed curiosity-based algorithm) is able to maximize the reward over time. In other words, only PPO+HCM can learn to perform according to the goals mentioned in Section \ref{environment_section} in the environment. The right side of the same figure displays some information about Failed Fly. A `Failed Fly' is when the quad-copter is initiated randomly in the environment, the algorithm cannot learn to control it and, as a result, the quad-copter crashes. Again, only PPO+HCM can control the quad-copter and reduce the number of `Failed Fly' over time, which is another sign that the algorithm is successful in the environment.
\par
Considering the problem our algorithm tries to solve is controlling a quad-copter, the averaged position and attitude errors are collected and displayed in Figure \ref{averaged_errors_fig}. The information shown in the figure can be divided into two categories 1) Attitude Control Information, and 2) Position Control Information. Reducing errors for Attitude Control implies that the algorithm can control the quad-copter and not crash. However, it does not give any information about how far or near the quad-copter is from the desired position. That information can be retrieved from the Position Control-related diagrams (i.e., the left side of the Figure \ref{averaged_errors_fig}). Overall, both Attitude and Position control information illustrates the capability of our proposed curiosity-based method (i.e., HCM) while showing the lack of ability of other algorithms in terms of control and flying toward the desired position and attitude.
\par
So far, our result showed this paper's algorithm capability in controlling the quad-copter agent and flying toward the desired position and attitude (i.e., the first part of the objectives of the environment described in Section \ref{environment_section}). However, to illustrate the algorithm's capabilities, some information regarding the obstacles is necessary (i.e., the second part of the environment objectives), which is mentioned in the following.

\subsection{Goal and Obstacles Distances}
As described in Section \ref{environment_section}, the algorithm should control a quad-copter agent and fly it toward a desired location while avoiding three obstacles. Figure \ref{averaged_distances_goal_obstacle_fig} is comprised of two diagrams. The left diagram shows the distance between the quad-copter and the position of the goal, and the right image shows the distance between the quad-copter and obstacles in the environment. The obstacle distance is the mean of the three Euclidean distances between the quad-copter and each one of the obstacles in the environment. Both diagrams in Figure \ref{averaged_distances_goal_obstacle_fig} show the capability of PPO+HCI in terms of decreasing the distance between the quad-copter and the location of the goal while slightly increasing and then maintaining the distance between the quad-copter and obstacles. These two diagrams illustrate that the quad-copter controlled by PPO+HCP reaches the goal while avoiding the obstacles.

\subsection{Curiosity Visualization}
Considering the description in Section \ref{curiosity_visualization_section} regarding the visualization of the curiosity effect, Figure \ref{curiosity_visualization_fig} and \ref{curiosity_visualization_fig_inc} show the effect of curiosity in evolving exploration patterns when the number of episodes increases. The figures are comprised of two rows; the first row shows the PPO exploration pattern, and the second row shows the PPO+HCM exploration pattern. For PPO, as can be seen in the figures, the pattern does not evolve and is almost static where it does not move toward the center of the box (i.e., toward the desired position); this observation is also supported by the Position Control Information section of Figure \ref{averaged_errors_fig}, as the algorithm does not reduce the position error by time. For the PPO+HCM, on the other hand, as the illustrated in the second row of the figures, the pattern of exploration is changing. By increment of the number of episodes, the algorithm explores more trajectories that ends up towards the center of the box (i.e., desired position), and also, more concentration can be seen in that area.

\subsection{Trajectories of Optimal Policy}
Figure \ref{optimal_policy_trajectories_fig} shows two sample trajectories generated by an optimal policy trained using the curiosity-based algorithm proposed in this paper (i.e., PPO+HCM) when the quad-copter is initiated in random initial positions and attitudes and three obstacles are initiated in three random positions. The trajectories are visualized using the right-hand coordinate system where the red arrow shows the Yaw orientation of the quad-copter. As can be seen, the algorithm can control the quad-copter and pass the obstacles while controlling the vehicle's Yaw toward the desired location.

\section{Discussion}
In this section, we discuss some matters that need further explanation.

\subsection{Algorithm Selection Strategy}
We used the PPO as the baseline algorithm for the performance test because it is the algorithm that is mostly used for learning to fly such as in \cite{RAMEZANIDOORAKI2021103671} \cite{Hwangbo_2017} \cite{MolchanovRL_LowLevel} and is a powerful on-policy algorithm. Moreover, we also selected SAC to show the result of a powerful off-policy RL algorithm that incorporates a memory replay. Our goal in this paper was to show that integrating curiosity with reinforcement learning-based flight controller makes solving complex low-level flight control problems possible. Thus, we selected PPO+ICM for testing the performance. Finally, we showed the capability of the proposed algorithm (i.e., PPO+HCM) compared with the competing ones and the importance of our contribution at the algorithm level.

\subsection{The Gap between the Simulation and Real World}
One area for improvement with the simulation-based learning algorithm is the gap between the simulation and the real world. One way to address this problem is by reducing the difference between the simulation and the real-world environments. However, this paper focuses on the learning part and the effect of adding curiosity to the learning policy for low-level flight control. For this reason, we did not enter the area of real-world tests in this paper, especially considering that it is already proven in \cite{Hwangbo_2017} \cite{MolchanovRL_LowLevel} \cite{ChenHuanPi_LowLevel} that a reinforcement learning-based policy can be used in a real-world quad-copter for low-level control.

\subsection{Exploration, Exploitation, and Curiosity}
It is a well-known capability of curiosity to increase and orient the exploration towards surprise, or in better words, to explore the environment meaningfully. While reinforcement learning algorithms, by default, use a trade-off between exploitation and random exploration (exploration mechanisms such as epsilon greedy in Q-learning or noise injected to the output action in methods such as DDPG, SAC, and PPO), they do purely random explorations. However, curiosity, as implemented in our work, generates an intrinsic reward that increases in the states unknown to the agent (by measuring the agent prediction error on those states) and decreases in the states that the agent visited frequently. Thus, curiosity can be seen as a parameterized neural network architecture that rewards the agent more in surprising states, motivating the agent to explore those states more. As a result, curiosity aims to make the exploitation part of the RL more efficient while still using the default exploration mechanism.

\subsection{Computational Time}
Considering adding an extra head to PPO networks and having ten sub-modules in the proposed curiosity algorithm, the learning part is much heavier than regular RL-based low-level flight control learning. However, that is only for the learning time, which usually happens on a powerful machine. For the execution time or deployment on an actual quad-copter, our approach is similar to the mentioned approaches because only one network (i.e.,  Policy Network) is needed.

\section{Conclusion}
\label{sec:conclusion}
In this work, we proposed a new approach for autonomous learning of low-level control policy. To achieve that, we proposed a new approach for implementing a computational model of curiosity motive using prediction error. To measure the capability of our algorithm, we designed and implemented a complex environment in Gazebo for learning quad-copter low-level flight control policy. In the designed environment, the algorithm should learn to directly control the quad-copter by generating the proper motor speed from odometry data. Further, the algorithm should learn to fly through obstacles while controlling the Yaw direction of the quad-copter toward the desired location. We ran tests to measure and compare the proposed algorithm to other baseline algorithms. 
\par
As shown in Section \ref{evaluation_section} of this paper, the proposed approach (i.e., PPO+HCM) can learn a flight policy when other algorithms fail to do so. By incorporating the proposed curiosity module, the algorithm can evolve the exploration pattern and fly to areas where other algorithms avoid flying. As a result, it can learn to control the quad-copter, control the Yaw direction of the quad-copter toward the desired location, and avoid hitting obstacles (as a low-level controller). 
%\par
In future works, we plan to measure the effect of incorporating machine imagination in a low-level framework.
%\appendix
%\section{Projection and Back-Projection}
%\input{sections/projection}

\section*{Acknowledgment}
This work is supported by the European Union’s Horizon 2020 Research and Innovation Program (OpenDR) under Grant 871449. This publication reflects the authors’ views only. The European Commission is not responsible for any use that may be made of the information it contains.
\bibliographystyle{IEEEtran}
\bibliography{CDRL_LLFC}

\end{document}